\documentclass[a4paper,conference]{IEEEtran}

\usepackage{cite}
\usepackage{amsmath,amssymb,amsfonts}
\usepackage{algorithmic}
\usepackage{graphicx}
\usepackage{textcomp}

\usepackage[utf8]{inputenc} 
\usepackage[T1]{fontenc}    
\usepackage{hyperref}       
\usepackage{url}            
\usepackage{booktabs}       
\usepackage{nicefrac}       
\usepackage{microtype}      
\usepackage{xcolor}         
\usepackage{bm}
\usepackage{bbm}
\usepackage{subcaption}
\usepackage{placeins}
\newcommand{\norm}[1]{\left\lVert#1\right\rVert_2}

\newcommand{\ra}[1]{\renewcommand{\arraystretch}{#1}}

\ifCLASSINFOpdf
\else
\fi
\begin{document}
%
\title{Exploiting Diversity of Unlabeled Data for Label-Efficient Semi-Supervised Active Learning}

\author{\IEEEauthorblockN{Felix Buchert, Nassir Navab}
\IEEEauthorblockA{Chair for Computer Aided Medical Procedures\\ Technical University of Munich\\
Garching bei München, 85748, Germany\\ Email: felix.buchert@tum.de, nassir.navab@tum.de}
\and
\IEEEauthorblockN{Seong Tae Kim}
\IEEEauthorblockA{Department of Computer Science and Engineering\\ Kyung Hee University\\
Yongin-si, 17104, South Korea \\ Email: st.kim@khu.ac.kr}}


\maketitle

\begin{abstract}
The availability of large labeled datasets is the key component for the success of deep learning. However, annotating labels on large datasets is generally time-consuming and expensive. Active learning is a research area that addresses the issues of expensive labeling by selecting the most important samples for labeling. Diversity-based sampling algorithms are known as integral components of representation-based approaches for active learning. In this paper, we introduce a new diversity-based initial dataset selection algorithm to select the most informative set of samples for initial labeling in the active learning setting. Self-supervised representation learning is used to consider the diversity of samples in the initial dataset selection algorithm. Also, we propose a novel active learning query strategy, which uses diversity-based sampling on consistency-based embeddings. By considering the consistency information with the diversity in the consistency-based embedding scheme, the proposed method could select more informative samples for labeling in the semi-supervised learning setting. Comparative experiments show that the proposed method achieves compelling results on CIFAR-10 and Caltech-101 datasets compared with previous active learning approaches by utilizing the diversity of unlabeled data.
\end{abstract}
\IEEEpeerreviewmaketitle

\section{Introduction}
Access to large labeled datasets has been a key factor in the successful training of deep neural networks. While unlabeled data is often abundantly available, human label acquisition is generally time-consuming and expensive. In particular, in application domains, in which the label acquisition has to be conducted by experts, such as the medical domain, the cost of building a large labeled dataset may become prohibitive and ultimately limit the development and widespread application of deep learning. \emph{Active learning} and \emph{semi-supervised learning} are two approaches that address this issue and their combination, called \emph{semi-supervised active learning}, has been leveraged to perform label efficient training of deep neural networks.

The main problem of active learning algorithms is the query strategy. Query strategies assess the informativeness of unlabeled samples and aim at selecting only the most informative samples for labeling at every active learning cycle. Representation-based active learning algorithms employ query strategies that aim at selecting samples that encode diverse representational information. In this context, diversity-based sampling algorithms such as the \emph{k-means++ initialization step} \cite{david2007kmeans++, ash2020badge} and the \emph{k-Center-Greedy algorithm} \cite{sener2017coreset} have been commonly used to select a diverse set of samples for labeling based on a given embedding space.

In this paper, we propose a diversity-based initial dataset selection algorithm based on self-supervised learning and a query strategy based on consistency-based embeddings, that is specifically designed for semi-supervised active learning.
Our contributions are summarized as follows:
\begin{enumerate}
    \item We introduce a novel method based on self-supervised learning methods to select informative initial labeled datasets for semi-supervised active learning. By selecting a set of informative samples for initial labeling based on their diversity, the proposed approach outperforms random selection, which is widely assumed in active learning methods.
    \item We propose a novel active learning query strategy, which aims to select the most informative samples for labeling in the semi-supervised active learning setting. The main idea is to use an embedding space that naturally encodes both a sample's representational information and the consistency of model predictions on it. Experimental results show that the balancing of these two information is important for selecting useful samples in the semi-supervised active learning setting.
\end{enumerate}

\section{Related Work}
\subsection{Active Learning} Recently deep batch active learning algorithms mostly follow one of two well-established approaches, namely uncertainty-based and representation-based approaches. Uncertainty-based active learning algorithms \cite{joshi2009entropy,scheffer2001margin, tong2001support, roth2006margin,gal2016dropout,yoo2019learningloss} aim at selecting samples for labeling on which the current model's predictions are highly uncertain. Representation-based active learning algorithms aim at selecting a diverse batch of samples, which encode diverse semantic and representational information and, ideally, are representative of the entire unlabeled dataset \cite{sener2017coreset,sinha2019vaal}. 

There are also methods that combine uncertainty and representativeness for sample selection. Ash et al. \cite{ash2020badge} introduce a method to embed the unlabeled dataset into a gradient embedding space, in which sample representations naturally encode both diversity and uncertainty. The initialization step of the \emph{k-means++} clustering algorithm is then used sample a diverse and uncertain batch of samples from the gradient embedding space.

\subsection{Semi-supervised Learning} Semi-supervised learning algorithms use both labeled and unlabeled data for model training. By extracting valuable information from a large pool of unlabeled data, they seek to improve predictions of a machine learning model and can significantly reduce the number of labeled samples required to train a deep learning model. Recent studies in semi-supervised learning for deep neural neural networks have been based on variations of key concepts such as \emph{pseudo-labeling} \cite{lee2013pseudo} and \emph{consistency regularization} \cite{sajjadi2016consistency_regularization}. On the basis of these concepts, \cite{tarvainen2017meanteacher,berthelot2019mixmatch,berthelot2019remixmatch,sohn2020fixmatch} have been proposed as deep semi-supervised learning algorithms.
\emph{Mean Teacher} \cite{tarvainen2017meanteacher} uses consistency regularization based on a so-called teacher model, which is constructed as an exponential moving average of model weights over previous training iterations. \emph{MixMatch} \cite{berthelot2019mixmatch} and \emph{ReMixMatch} \cite{berthelot2019remixmatch} employ the augmentation strategy \emph{MixUp} \cite{zhang2018mixup} and combine both pseudo-labeling and consistency regularization in a sophisticated manner. By contrast, \emph{FixMatch} \cite{sohn2020fixmatch} combines pseudo-labeling and consistency regularization in a simplistic yet powerful manner. 

\subsection{Self-supervised Learning}
Unsupervised representation learning is a field of machine learning, which aims at learning (usually low-dimensional) representations of data samples that facilitate the extraction of useful information \cite{bengio2013representation}. Self-supervised learning is widely used to learn low-dimensional representations of data samples that facilitate the extraction of useful information \cite{bengio2013representation, gidaris2018unsupervised, grill2020BYOL, chen2020simCLR, pathak2016context, chen2020simple}. It relies on the definition of a pretext task and only requires a sufficiently large, unlabeled image dataset. Effective pretext tasks force the model to learn semantic features of input images, which have been shown to generalize well to other computer vision tasks. \emph{RotNet} \cite{gidaris2018unsupervised}, \emph{context encoders} \cite{pathak2016context}, \emph{SimCLR} \cite{chen2020simCLR} and \emph{BYOL} \cite{grill2020BYOL} are examples of successful self-supervised representation learning algorithms. Ideally, the learned representations encode characteristic sample features which can be used effectively in downstream tasks.

\section{Diversity-based Sampling in Active Learning}
\label{sec:div_sampling}
Diversity-based sampling algorithms have been used in active learning to select an informative set of samples based on their representations in a given embedding space. For instance, the \emph{k-means++ initialization step} \cite{ash2020badge} and the \emph{k-Center-Greedy algorithm} \cite{sener2017coreset} are commonly used in this context. In the following, we review formulations for diversity-based sample selection. We further discuss how specifically constructed embedding spaces can be used to balance multiple sample characteristics, which is of particular interest in semi-supervised active learning. 

Let $\mathcal{C}$ denote the set of selected samples, which is initialized to the empty set at the beginning of the algorithms. In general, diversity-based sampling algorithms select samples based on their distance to the closest selected sample embedding in an iterative way. More formally, for the $i$-th sample, this distance measure is given by
\begin{equation}
    d_i = \min_{j \in \mathcal{C}} \norm{\boldsymbol{z}_i - \boldsymbol{z}_j},
\end{equation}
where $d_i$ denotes the $L_2$-distance between the sample embedding $\boldsymbol{z}_i$ and the embedding of the closest (already selected) sample in $\mathcal{C}$. Samples are selected according to the categorical probability distribution given by
\begin{equation}
    p\left(\boldsymbol{z} = \boldsymbol{z}_i\right) = \frac{d_i^{\frac{1}{T}}}{\sum_{j=1}^N d_j^{\frac{1}{T}}}
    \label{equation:generalized_sampling}
\end{equation}
where $p\left(\boldsymbol{z} = \boldsymbol{z}_i\right)$ denotes the probability of selecting the sample with index $i$ and $T$ is a hyperparameter. After every sampling step, the set $\mathcal{C}$ and the distances $d_i$ are updated prior to selecting the next sample. This iterative process is repeated until the desired number of samples has been selected. \emph{k-Center-Greedy algorithm} and \emph{k-means++ initialization step} are represented in Eq. \ref{equation:generalized_sampling}.

\begin{figure*}[!t]
    \centering
    \includegraphics[width=1\textwidth]{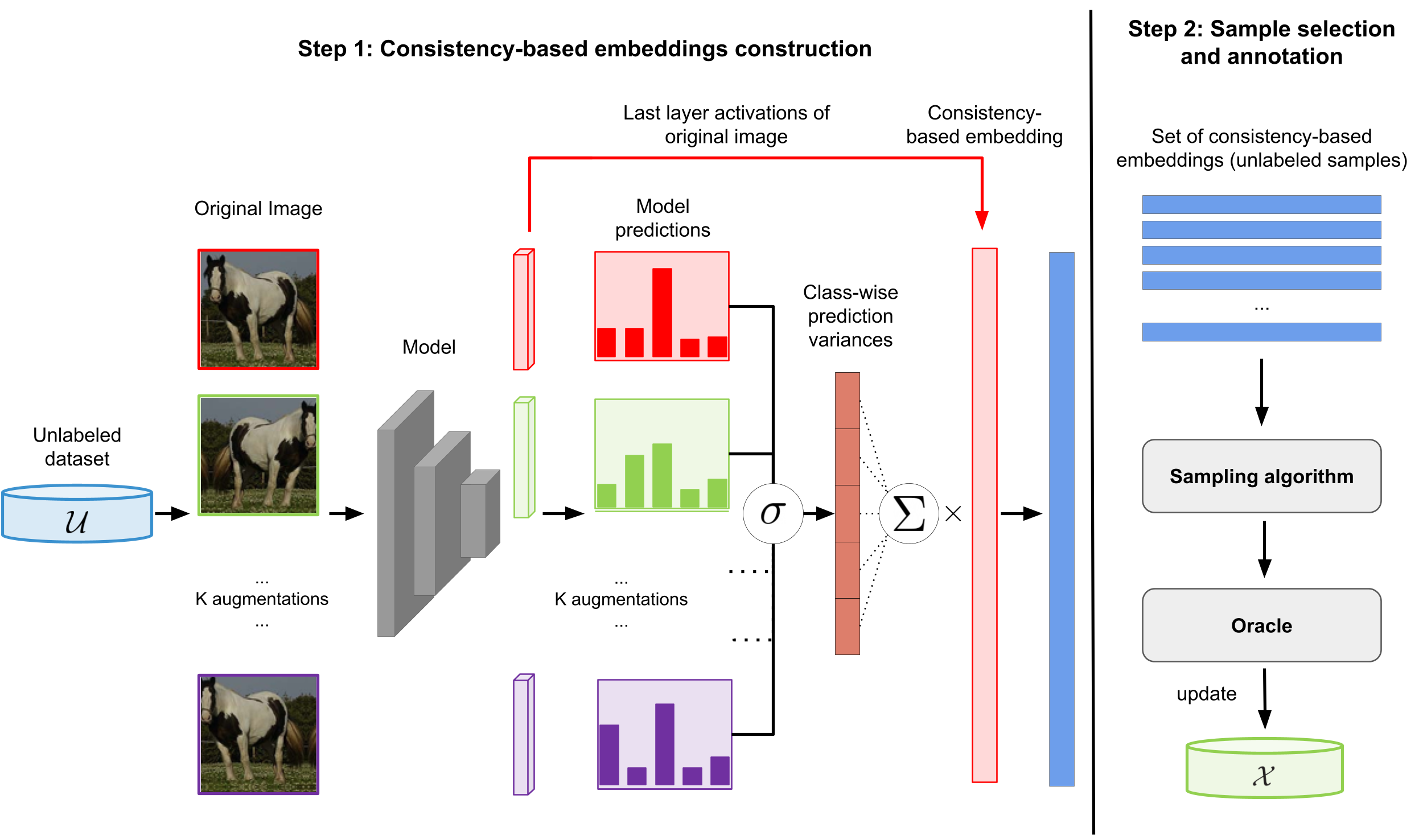}
    \caption{Overview of consistency-based embeddings query strategy pipeline. The overall pipeline largely consists of two steps: 1) Consistency-based embedding construction, 2) sample selection and annotation.}
    \label{fig:consistency_embedding_overview}
\end{figure*}
\section{Diversity-based Sampling in Semi-Supervised Active Learning}
\label{sec:div_sampling_ssl_al}
We introduce two interesting diversity-based sampling approaches in the context of semi-supervised active learning. Firstly, we discuss diversity-based sampling for initial dataset selection using representations of self-supervised learning. Secondly, we introduce a novel active learning query strategy, which is based on diversity-based sampling from the consistency-based embedding space.

\subsection{Diversity-based Sampling for Initial Dataset Selection}
\label{subsec:initial_sampling}
Semi-supervised active learning algorithms generally start from a small labeled dataset and iteratively query labels for additional samples over multiple active learning cycles. Hence, algorithms that aim at selecting the most informative \emph{initial datasets} for labeling can potentially improve the performance of semi-supervised active learning in subsequent cycles. 

Active learning research has shown that it is possible to assess the informativeness of a batch of samples. One effective and robust approach is to consider the diversity of representational information in a batch of samples \cite{sener2017coreset}. 
Contrary to query strategies in active learning, there is no trained model in the initial dataset selection. The key challenge is to assess the diversity of representational information without having access to any label information at all. The advances in self-supervised learning \cite{gidaris2018unsupervised, chen2020simCLR, grill2020BYOL} have shown that it is possible to learn semantically meaningful representations from purely unlabeled data, which motivates our approach to initial dataset selection. It consists of two basic steps: a representation learning step and a diversity-based sampling step. In the first step, a self-supervised learning is used to generate embedding vectors for each sample in the pool of unlabeled data. Depending on the size and complexity of the given unlabeled dataset, different representation learning algorithms can be used in this step. \autoref{sec:results} presents both empirical and qualitative analysis of different representation learning algorithms \cite{gidaris2018unsupervised, grill2020BYOL} applied to different datasets.
In the second step, we aims at selecting an informative set of initial samples for labeling based on the embeddings generated in the representation learning step. For this purpose, we use the diversity-based sampling to select a diverse set of samples, which, ideally, is representative of the entire unlabeled dataset. On the basis of the empirical analysis, this study uses the hyperparameter setting $T=0.5$, i.e. the kmeans++ initialization step, to select informative initial samples for labeling.

\subsection{Diversity-based Sampling of Consistency-based Embeddings}
\label{subsec:cons_query_strategy}
In addition to diversity-based sampling in the context of initial dataset selection, we explore its application in the context of the semi-supervised active learning query strategy.

Gao et al. \cite{gao2020consistencySSLAL} have proposed using \emph{consistency}, referring to the consistency of model predictions on augmented versions of a given unlabeled sample, as selection criterion. The central assumption is that for unlabeled samples with highly inconsistent predictions, the semi-supervised learning algorithm presumably has not succeeded at effectively using them for model training. This assumption is motivated by the fact that recent semi-supervised learning algorithms \cite{berthelot2019mixmatch, berthelot2019remixmatch, sohn2020fixmatch} use the concept of consistency regularization. If model predictions on a given sample are still inconsistent after model training, it is reasonable to assume that the semi-supervised learning algorithm cannot extract useful information from that sample and therefore labeling on this sample can be highly informative.  

However, the purely consistency-based approach \cite{gao2020consistencySSLAL} does not explicitly consider the diversity of the selected samples. As discussed in \cite{sener2017coreset, ash2020badge, kirsch2019batchbald}, this can lead to the selection of samples with high overlap in representational information, which can result in many redundant samples in a selected batch. To address this issue, we propose an active learning method that combines diversity and consistency for selecting samples in semi-supervised learning setting. By applying diversity-based sampling to an embedding space specifically designed for semi-supervised active learning, our method aims at balancing both consistency and diversity as selection criteria.

Let $\bm{\mathcal{U}}_t = \{\boldsymbol{u}_i: i \in \left(1, \dots, N_t\right)\}$ denote the pool of unlabeled samples and $M_t$ denote the trained target model at active learning step $t$. We characterize the consistency of model $M_t$ on an unlabeled sample $\boldsymbol{u}_i$ by evaluating class-wise prediction variances $\sigma_{i,c}^2$ on $K$ augmented versions of it, denoted by $ \Tilde{\boldsymbol{u}}_{i,k} = \alpha\left(\boldsymbol{u}_i\right)$, as follows:
\begin{equation}
    \begin{aligned}
        \sigma_{i,c}^2 &= \mathrm{Var}\left[p_{M}\left(y_c | \boldsymbol{u}_{i,1}\right), p_M\left(y_c | \Tilde{\boldsymbol{u}}_{i,1}\right), \dots, p_M\left(y_c | \Tilde{\boldsymbol{u}}_{i,K}\right) \right]
    \end{aligned}
    \label{eq:pred_variance}
\end{equation}
$\alpha\left(\cdot\right)$ denotes a standard augmentation operation, $\mathrm{Var}[\cdot]$ denotes the sample variance and $\boldsymbol{\sigma}_i^2 = \left(\sigma_{i,1}^2, \dots, \sigma_{i, C}^2\right)$ 
is the vector in which the c-th element is given by the sample variance of predictions for class $c \in \{1, \dots, C\}$.

Given a sample $\boldsymbol{u}_i$, the activations of the penultimate layer of the target model $M_t$, denoted by $\boldsymbol{v}_i$, encode sample-specific representational information. On the basis of the embeddings $\boldsymbol{v}_i$ and the vector of class-wise prediction variances $\boldsymbol{\sigma}_i^2$, we define a consistency-based embedding $\boldsymbol{z}_i$ for a sample $\boldsymbol{u}_i$ as the last-layer activations $\boldsymbol{v}_i$ scaled by the sum of class-wise prediction variances, i.e.
\begin{equation}
    \boldsymbol{z}_i = \left(\sum_{c=1}^C\sigma_{i,c}^2\right) \cdot \boldsymbol{v}_i
    \quad\mathrm{and}\quad
    \norm{\boldsymbol{z}_i} = \left(\sum_{c=1}^{C}\sigma_{i,c}^2\right) \cdot \norm{\boldsymbol{v}_i}.    \label{eq:agg_consistency_embedding}
\end{equation}
The consistency-based embeddings have the same dimensionality as the last-layer activations, i.e. $\boldsymbol{z}_i \in \mathbb{R}^{d}$. By construction, the norm of the consistency-based embeddings is proportional to the sum of class-wise prediction variances. Subsequent to the construction of consistency-based embeddings, a diversity-based sampling algorithm is used to select samples. \autoref{fig:consistency_embedding_overview} shows the pipeline of the proposed consistency-based embedding query strategy.
Empirically, diversity-based algorithms have been shown to select diverse and high-magnitude embeddings. Hence, given the consistency-based embeddings, the sampling algorithm is ought to select a batch of samples, which is both diverse and on which the model's predictions are inconsistent (i.e. large embedding norms). This demonstrates how diversity-based sampling algorithms can be used to naturally balance selection criteria when applied to a suitable embedding space.

\section{Experiments and Results}
\label{sec:results}
\subsection{Experimental Settings and Implementation}
\label{subsec:settings}
The experimental results presented in this section have been collected on the basis of the general experimental setup described in the following.

\vspace{9mm}
\begin{table*}[!t]
\centering
\scalebox{1}{
\ra{1.3}
\begin{tabular}{@{}lcc@{}}
    \toprule
    & CIFAR-10 & Caltech-101\\ 
    \midrule
    Network architecture & Wide ResNet-28-2 & ResNet-18\\
    Active learning configuration\\
    \hspace{5mm}$Representation\:learning\:algorithm$ & RotNet \cite{gidaris2018unsupervised} & BYOL \cite{grill2020BYOL}\\
    \hspace{5mm}$Initial\:dataset\:size$ & 150 (0.3\%) & 388 (5.0\%)\\
    \hspace{5mm}$Budget\:sizes$ & [50, 50, 250, 250, 250] & [388, 388, 388]\\
    \hspace{5mm}$Independent\:runs$ & 5 & 3\\
    \hspace{5mm}$No.\:of\:augmentations$ & 50 & 10\\
    MixMatch settings (semi-supervised learning)\\
    \hspace{5mm}$Epochs$ & 1024 & 64\\
    \hspace{5mm}$Finetuning\:epochs$ & 128 & 16\\
    \hspace{5mm}$\lambda_U\:(Unlabeled\;loss\;weight)$ & 75 & 150\\
    \bottomrule
    \end{tabular}}
    \caption{Overview of experiment settings including network architectures, active learning configuration, and semi-supervised learning setting for the proposed method on CIFAR-10 and Caltech-101 datasets.}
\label{tab:experiment_settings}
\end{table*}
\subsubsection{Datasets} All experiments are conducted on two public image classification datasets, CIFAR-10 \cite{krizhevsky2009cifar} and Caltech-101 \cite{fei2004caltech101}. The CIFAR-10 dataset consists of 60,000 images, which belong to ten different classes. The images are split into a training set of 50,000 images and a test set of the remaining 10,000 images.
The Caltech-101 dataset consists of a total of 8,677 images belonging 101 different classes. There are between 40 and 800 images per class. As is common practice, the Caltech-101 images are resized to a size of 224$\times$224 for all experiments. We randomly split it into a training set consisting of 90\% of the entire dataset and a test set constituted of the remaining 10\%.
\subsubsection{Implementation}
Following \cite{gao2020consistencySSLAL}, MixMatch \cite{berthelot2019mixmatch} is used for model training in this study. In accordance with the setting presented in \cite{song2019mixmatchAL}, a Wide ResNet-28-2 \cite{zagoruyko2016wide_resnet} is used as network architecture for experiments on CIFAR-10. For experiments on Caltech-101, a ResNet-18 is used as network architecture. All active learning algorithms are run with the same learning rate of 0.002 and weight decay of 0.00004 (as in \cite{berthelot2019mixmatch}) using the Adam optimizer \cite{kingma2014adam}. The size of the initial labeled dataset and the budget sizes, i.e. the number of samples selected for labeling at every active learning cycle, are chosen as in \cite{gao2020consistencySSLAL}. In all experiments, the standard augmentation operation used for the calculation of class-wise prediction variances (see \autoref{eq:pred_variance}) is implemented as random horizontal flips and random crops. Class-wise prediction variances are computed based on model predictions on 50 random augmentations for CIFAR-10 \cite{gao2020consistencySSLAL} and 10 random augmentations on Caltech-101. 
Table \ref{tab:experiment_settings} summarizes the experimental settings on each dataset.

\subsubsection{Comparison} In this study, we compare our query strategy with \emph{maximum entropy}, \emph{Coreset}, \emph{BADGE} and \emph{consistency-based semi-supervised active learning}. Maximum entropy \cite{joshi2009entropy} serves as purely uncertainty-based baseline algorithm, which selects the unlabeled samples for which the model's predictions have the highest entropy at every active learning cycle. \emph{Coreset} \cite{sener2017coreset} aims on selecting a diverse batch of samples using the activations of the penultimate network layer as embedding and the $L_2$-distance between sample embeddings as measure of diversity. \emph{BADGE} \cite{ash2020badge} constructs gradient embeddings, which encode both diversity and uncertainty (expected model change). Subsequently, the k-means++ initialization algorithm is applied to choose a set of informative samples by balancing diversity and uncertainty. Maximum entropy, Coreset and BADGE use supervised learning for model training and start from randomly selected initial dataset. 
\emph{Consistency-based semi-supervised active learning} \cite{gao2020consistencySSLAL} (also referred to as ``Consistency'' in the following) uses the sum of class-wise prediction variances over weakly augmented versions of a given sample as selection criterion. It uses MixMatch \cite{berthelot2019mixmatch} as semi-supervised learning algorithm and randomly selects the initial dataset. Public implementations of these selection baselines are followed \cite{ash2020badge_github, sener2017coreset_github}.
For evaluation, we calculate test accuracy over the different active learning stages (different number of labeled training samples). 

\begin{figure}[t]
    \centering
    \begin{subfigure}{0.49\textwidth}
        \includegraphics[width=1.0\textwidth]{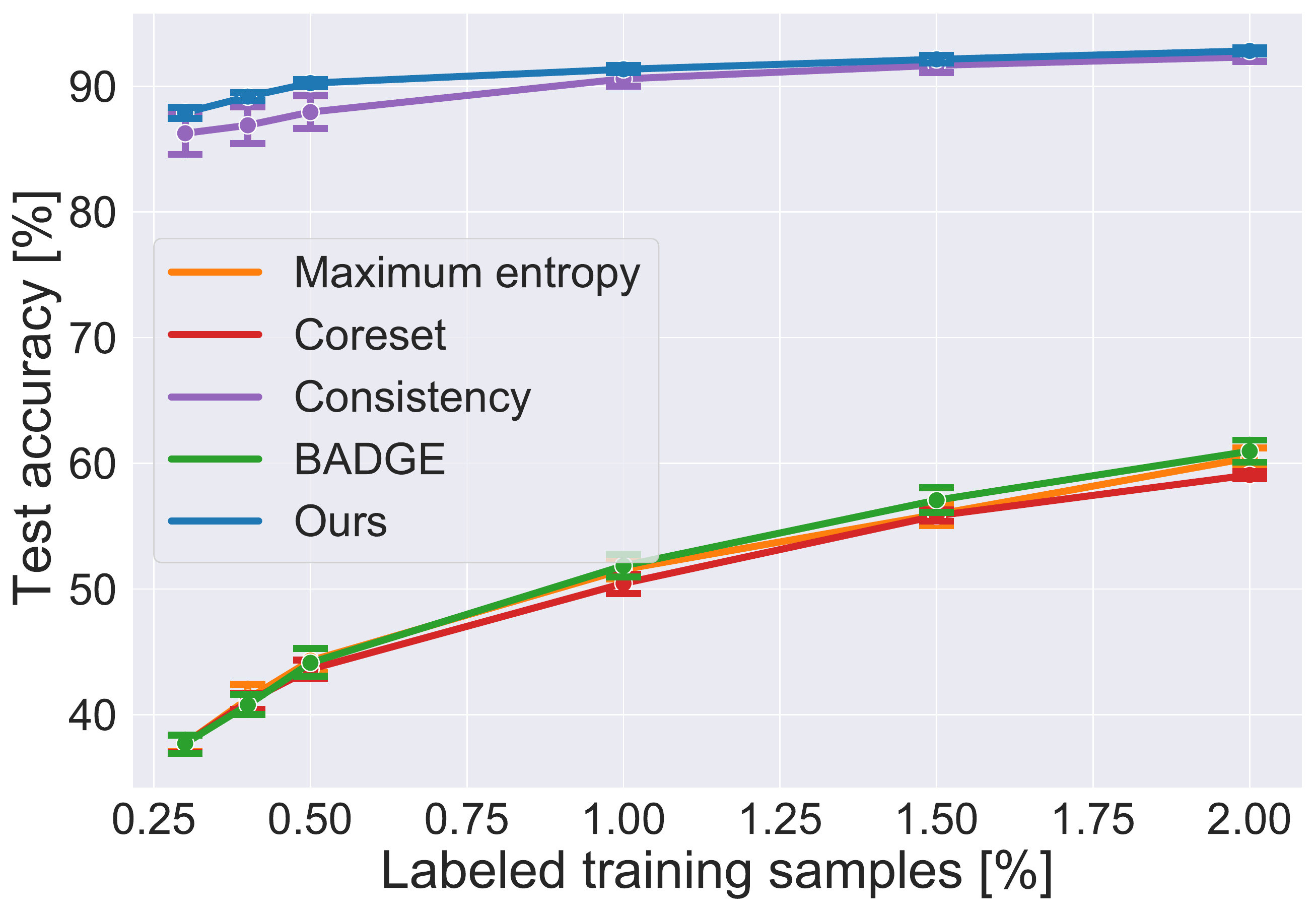}
        \caption{CIFAR-10}
        \label{cifar10}
    \end{subfigure}
    \begin{subfigure}{0.49\textwidth}
        \includegraphics[width=1.0\textwidth]{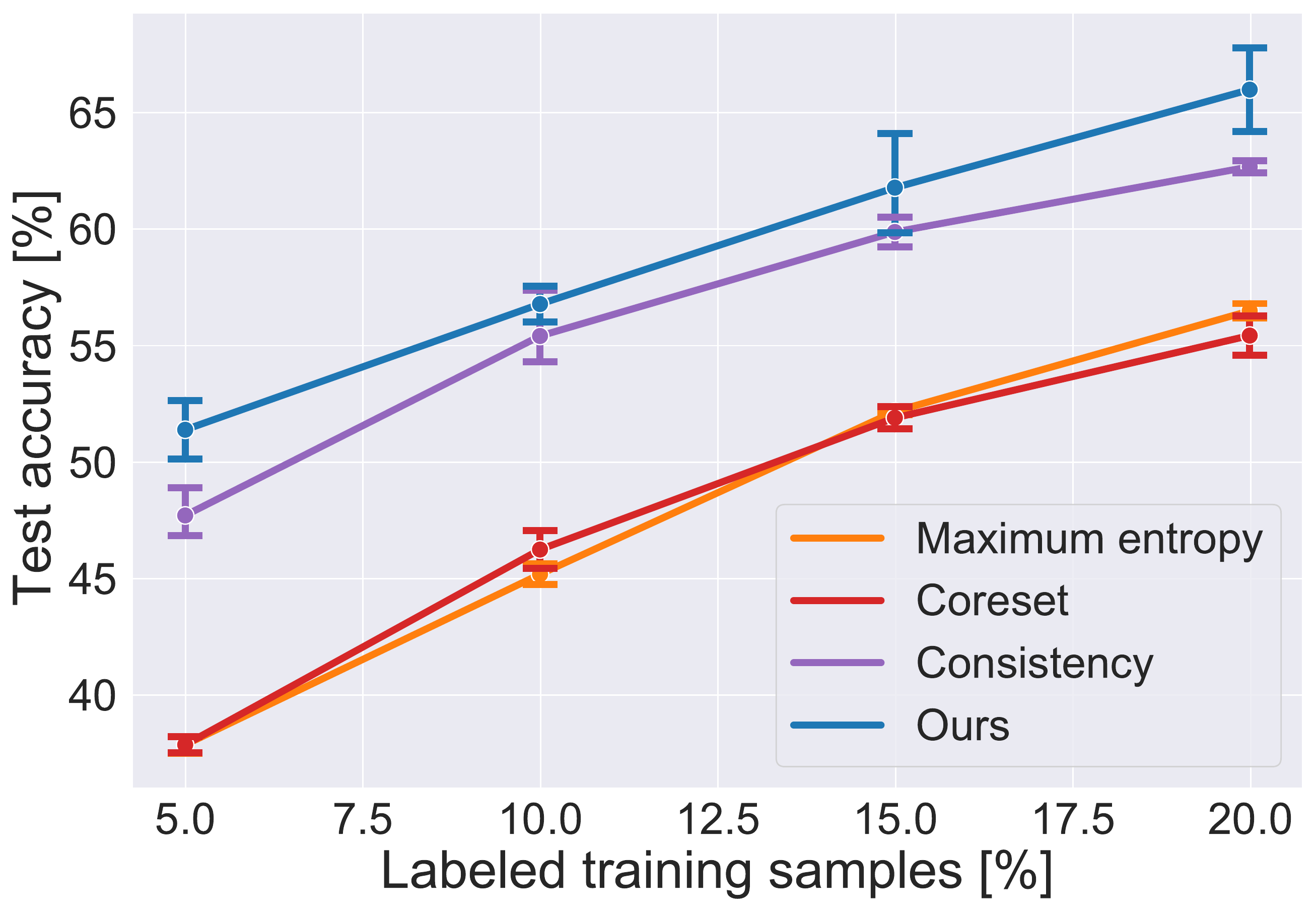}
        \caption{Caltech-101}
        \label{caltech101}
    \end{subfigure}
    \caption{Comparison of the proposed approach to baseline algorithms based in the active learning. The bars represent the standard error. 
    }
    \label{fig:pipeline_results}
\end{figure}
\subsection{Comparison of Performance in Active Learning}
\label{subsec:performance}
We examine the performance of our approach to semi-supervised active learning using both the initial dataset selection algorithm and the consistency-based embeddings query strategy. The baseline algorithms comprise Maximum entropy \cite{joshi2009entropy}, Coreset \cite{sener2017coreset} and BADGE \cite{ash2020badge} as well as purely consistency-based semi-supervised active learning (we name this as Consistency)\cite{gao2020consistencySSLAL}. While maximum entropy, Coreset and BADGE rely on supervised learning, consistency-based semi-supervised active learning and our algorithm perform semi-supervised learning using MixMatch \cite{berthelot2019mixmatch}. 
All baseline algorithms start from initial models trained on the same randomly selected initial datasets. As the computational complexity of BADGE scales linearly with the number of classes, it was only run on CIFAR-10 dataset. Due to the large cost of computation of semi-supervised active learning, we finetune models after every active learning cycle as in \cite{gao2020consistencySSLAL}. We follow the testing protocol employed in \cite{berthelot2019mixmatch, berthelot2019remixmatch, sohn2020fixmatch}. Accordingly, the test accuracy is computed using an exponential moving average of model parameters (decay rate of 0.999) at every epoch. Furthermore, we use the median accuracy of the last 20 epochs on CIFAR-10 \cite{berthelot2019mixmatch} and the median accuracy of the last 6 epochs on Caltech-101. Finally, we report the average accuracy over 5 runs with different random seeds. For initial dataset selection algorirthm in our approach, the embedding is provided by RotNet \cite{gidaris2018unsupervised} on CIFAR-10. For Caltech-101, a more challenging dataset, state-of-the-art self-supervised representation learning algorithm \emph{BYOL} \cite{grill2020BYOL} with a ResNet-50 pretrained on Imagnet \cite{yaox2020byol_github} is used to generate sample embeddings. 
In accordance with \cite{grill2020BYOL}, the activations of the penultimate network layer are used as sample representations resulting in embeddings of dimension 2024.

\autoref{fig:pipeline_results} shows the active learning results with respect to the test accuracy over multiple labeled training sample stages. The proposed approach outperforms all baseline algorithms over all active learning cycles on both datasets. It can be observed that algorithms based on semi-supervised learning outperform supervised active learning algorithms by a large margin, which highlights the general potential of combining semi-supervised learning and active learning. On CIFAR-10, the proposed approach achieves an accuracy of 92.81\% with 2\% of all samples labeled. On Caltech-101, it achieves an accuracy of 65.99\% with 20\% of all samples labeled. For reference, supervised active learning algorithms achieve accuracies of approximately 60\% on CIFAR-10 and 56\% on Caltech-101. 
The purely consistency-based semi-supervised learning algorithm \cite{gao2020consistencySSLAL} achieves an accuracy of 92.34\% on CIFAR-10 and 62.68\% on Caltech-101.

\begin{figure}[!t]
    \centering
    \begin{subfigure}{0.49\textwidth}
        \includegraphics[width=1.0\textwidth]{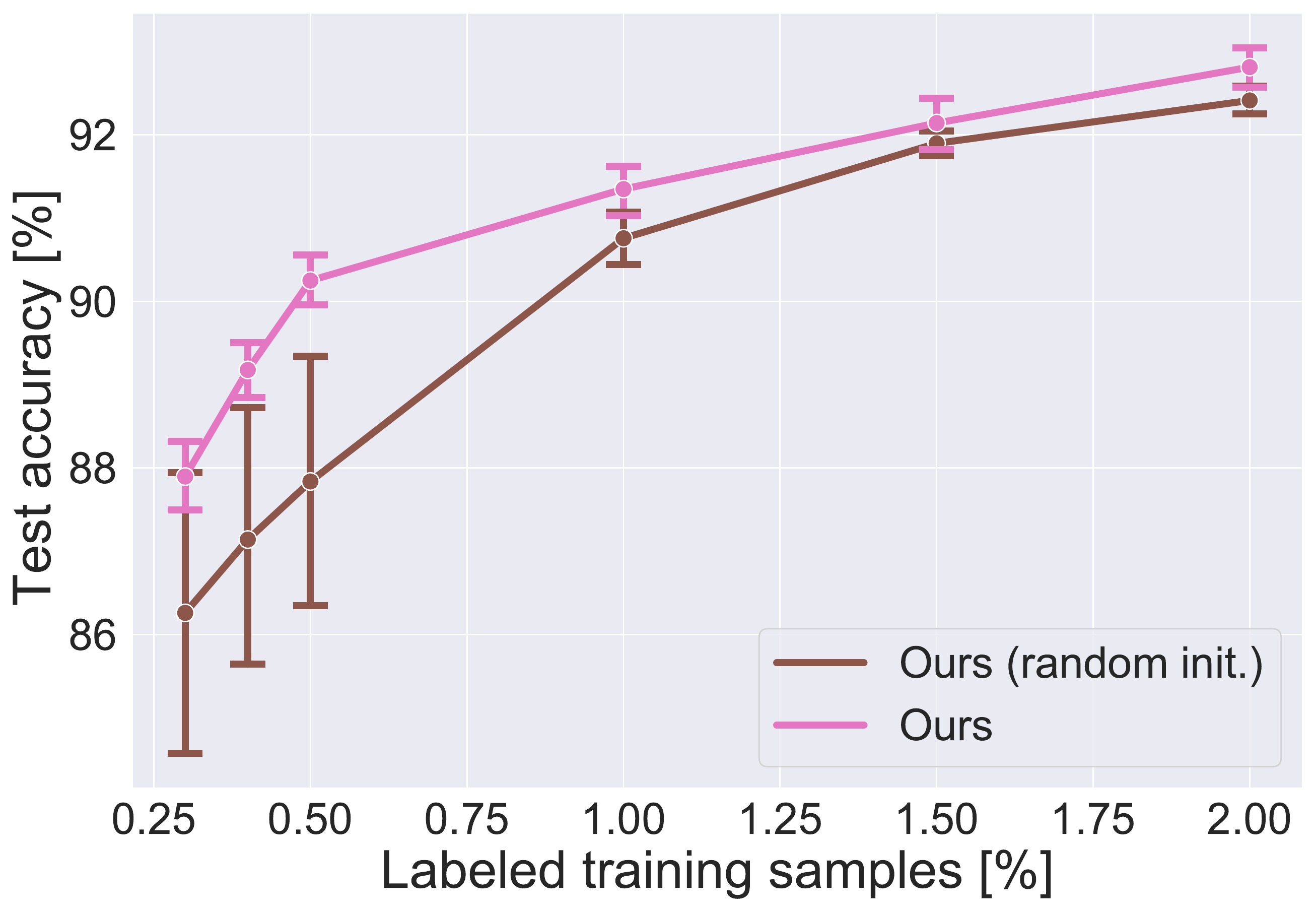}
        \caption{CIFAR-10}
        \label{fig:acc_init_sampling_cifar10}
    \end{subfigure}
    \begin{subfigure}{0.49\textwidth}
        \includegraphics[width=1.0\textwidth]{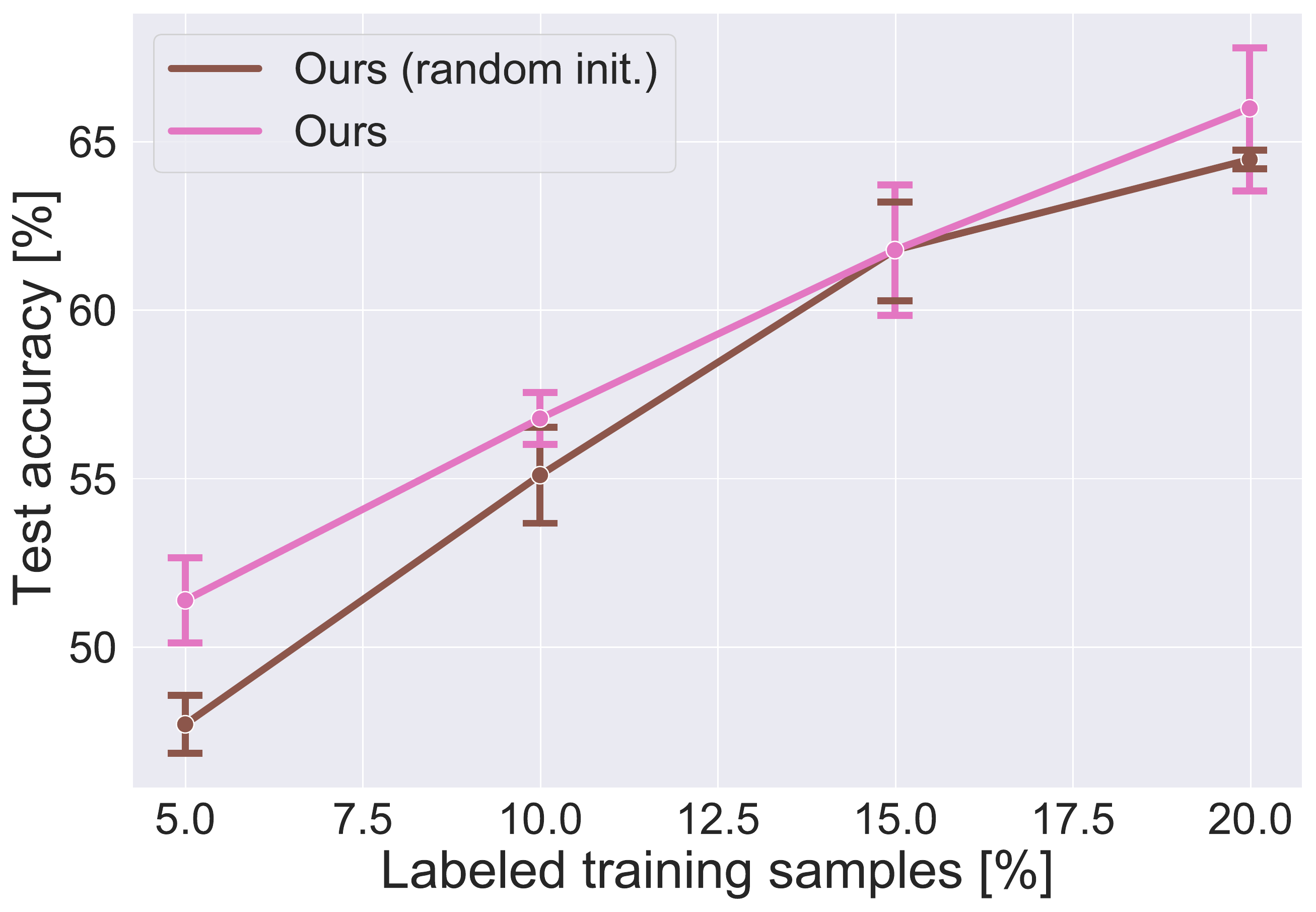}
        \caption{Caltech-101}
        \label{fig:acc_init_sampling_caltech101}
    \end{subfigure}
    \caption{Validation of the initial datasets selection algorithm based on test accuracy. The bars represent the standard error.}
    \label{fig:initial_sampling_validation}
\end{figure}
\subsection{Ablation Study}
We further provide empirical evidence validating the effectiveness of the initial dataset selection algorithm.
\autoref{fig:initial_sampling_validation} shows a comparison of the performance obtained in the two different settings on both CIFAR-10 and Caltech-101. One can observe that the proposed initial sample selection strategy achieves improvements of the accuracy of initial models. The accuracy of initial models is improved from 86.26\% to 87.90\% on CIFAR-10 and 47.71\% to 51.39\% on Caltech-101. In addition to that, the improvement persists throughout subsequent active learning cycles. Notably the accuracy on the last active learning step is increased from 92.41\% to 92.81\% on CIFAR-10 and from 64.47\% to 65.99\% on Caltech-101. 

\section{Discussion}
\label{sec:discussion}
The results presented above substantiate the effectiveness of the diversity-based sampling for evaluating and analyzing its applications in the context of semi-supervised active learning. In particular, we considered two core applications: an initial dataset selection algorithm and a novel query strategy for semi-supervised active learning.

We showed that the combination of both proposed components clearly outperforms all baseline algorithms on CIFAR-10 and Caltech-101. Notable gains in the performance of initial models on both CIFAR-10 and Caltech-101 demonstrate the effectiveness our diversity-based initial dataset selection algorithm. The consistency-based embeddings query strategy further improved the performance by explicitly considering diversity for sample selection. 
In particular, the application of diversity-based sampling to a specially constructed embedding space has proven to provide a good framework for naturally combining different selection criteria in semi-supervised active learning. Previous studies had either only considered selection criteria originally proposed for the supervised active learning setting \cite{song2019mixmatchAL, sener2017coreset} or solely focused on a single criterion \cite{gao2020consistencySSLAL}. By contrast, our method has proven to succeed at balancing both consistency and diversity as selection criteria.

It is clear that semi-supervised active learning algorithms are expected to significantly outperform supervised active learning. However, we think this comparison is still informative as it highlights the potential of semi-supervised active learning. Furthermore, \cite{gao2020consistencySSLAL} had already shown that their purely consistency-based approach to semi-supervised active learning outperforms straightforward combinations of semi-supervised learning and the considered baseline active learning algorithms.

Our findings aim at improving the label-efficiency for deep neural networks on image classification. This might be highly beneficial in medical applications, where the label acquisition process is particularly expensive and time-consuming. 

\section{Conclusion}
\label{sec:conclusion}
In this study, we introduced diversity-based sampling algorithms for semi-supervised active learning. The consistency-based embeddings query strategy highlighting that diversity-based sampling can be applied to a specifically constructed embedding space in order to naturally balance effective selection criteria. Furthermore, we demonstrate that the empirical analysis on both proposed components translates into gains of performance in active learning for image classification. We hope the presented concepts inspire future research to improve the label-efficiency of neural network training.

\section*{Acknowledgment}
This work was supported in part by the National Research Foundation of Korea(NRF) grant funded by the Korea government(MSIT) (No.2021R1G1A1094990), the Institute of Information and Communications Technology Planning and Evaluation (IITP) grant funded by the Korea Government (MSIT) (Artificial Intelligence Innovation Hub) under Grant 2021-0-02068, and a grant from Kyung Hee University in 2021 (KHU-20210732). S.T. Kim is a corresponding author (email: st.kim@khu.ac.kr). 

\bibliographystyle{IEEEtran}
\bibliography{bibliography}

\end{document}